\documentclass[11pt,a4paper]{article}
\usepackage[hyperref]{acl2019}
\usepackage{times}
\usepackage{latexsym}

\usepackage{url}
\usepackage{url}
\usepackage{graphicx}
\usepackage{wrapfig}
\usepackage{amsmath}
\usepackage{amssymb}
\usepackage{color,xcolor,colortbl}
\usepackage{enumitem}
\usepackage{algorithm}
\usepackage{algorithmic}
\usepackage[font={small}]{caption}
\usepackage{bm,bbm}
\usepackage{booktabs}
\usepackage{mathtools}
\usepackage{array}
\usepackage{subcaption}
\usepackage{pbox}
\usepackage{pifont}
\usepackage{multirow}
\usepackage{tikz}
\usepackage{tikz-cd}
\usepackage{xcolor}
\usetikzlibrary{arrows,arrows.meta}
\tikzcdset{arrow style=tikz, diagrams={>=stealth'}}

\DeclareMathOperator*{\argmax}{arg\,max\,}
\newcommand\numberthis{\addtocounter{equation}{1}\tag{\theequation}}

\definecolor{darkblue}{rgb}{0.0, 0.0, 0.55}
\newenvironment{fontppl}{\fontfamily{ppl}\selectfont}{\par} 
\definecolor{input}{rgb}{0.63, 0.79, 0.95}
\definecolor{conv1}{rgb}{1.0, 0.89, 0.77}
\definecolor{maxpool}{rgb}{0.92, 0.3, 0.26}
\definecolor{dense}{rgb}{0.82, 0.62, 0.91}
\definecolor{relu}{rgb}{0.76, 0.33, 0.76}

\usepackage{multirow}
\usepackage{comment}

\usepackage{soul}
\newcommand{\hlc}[2][yellow]{ {\sethlcolor{#1} \hl{#2}} }

\definecolor{1d}{HTML}{00BBFF}
\definecolor{2d}{HTML}{00BB00}
\definecolor{3d}{HTML}{F4D03F}
\definecolor{4d}{HTML}{DC6161}
\definecolor{5d}{HTML}{B27363}
\definecolor{6d}{HTML}{AF7AC5}
\definecolor{7d}{HTML}{9DA5AE}

\definecolor{1l}{HTML}{00FFFF}
\definecolor{2l}{HTML}{00FF00}
\definecolor{3l}{HTML}{FFFF00}
\definecolor{4l}{HTML}{FF9898}
\definecolor{5l}{HTML}{B9968D}
\definecolor{6l}{HTML}{D7BDE2}
\definecolor{7l}{HTML}{D6DBDF}

\aclfinalcopy 
\def\aclpaperid{2004} 

\title{Scoring Sentence Singletons and Pairs for Abstractive Summarization}

\author{Logan Lebanoff$^\dagger$ \quad Kaiqiang Song$^\dagger$ \quad Franck Dernoncourt$^\S$ \\
\textbf{Doo Soon Kim$^\S$ \quad Seokhwan Kim$^\S$ \quad Walter Chang$^\S$ \quad Fei Liu$^\dagger$}\\
\\
$^\dagger$Computer Science Department, University of Central Florida, Orlando, FL 32816 \\
\texttt{\footnotesize \{loganlebanoff, kqsong\}@knight.ucf.edu \quad feiliu@cs.ucf.edu}\\
\\[-0.5em]
$^\S$Adobe Research, San Jose, CA 95110 \\
\texttt{\footnotesize \{dernonco,dkim,seokim,wachang\}@adobe.com}
}

\date{}

\begin{document}
\maketitle
\begin{abstract}

When writing a summary, humans tend to choose content from one or two sentences and merge them into a single summary sentence.
However, the mechanisms behind the selection of \emph{one} or \emph{multiple} source sentences remain poorly understood.
Sentence fusion assumes multi-sentence input;
yet sentence selection methods only work with single sentences and not combinations of them.
There is thus a crucial gap between sentence selection and fusion to support summarizing by both compressing single sentences and fusing pairs.
This paper attempts to bridge the gap by ranking sentence singletons and pairs together in a unified space.
Our proposed framework attempts to model human methodology by selecting either a single sentence or a pair of sentences, then compressing or fusing the sentence(s) to produce a summary sentence.
We conduct extensive experiments on both single- and multi-document summarization datasets and report findings on sentence selection and abstraction.

\end{abstract}

\section{Introduction}
\label{sec:intro}

\begin{figure}[t]
\centering
\includegraphics[width=3in]{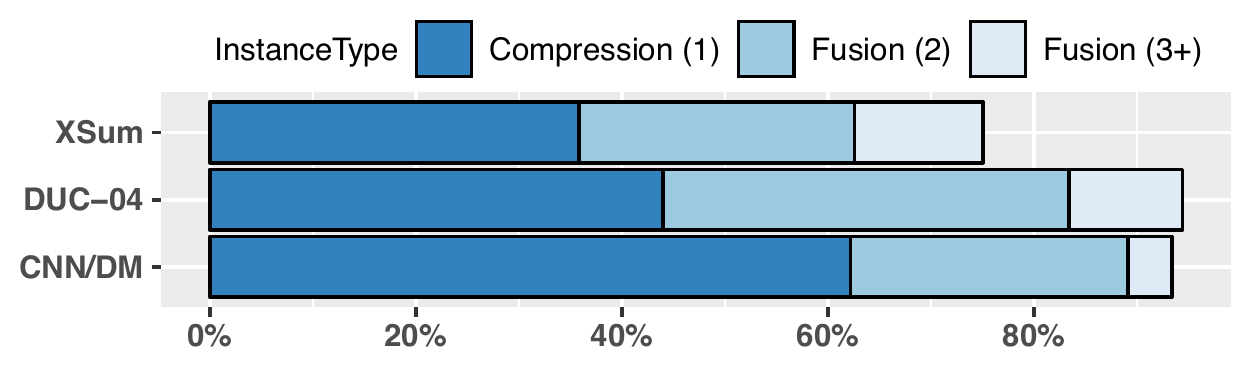}
\caption{Portions of summary sentences generated by compression (content is drawn from one source sentence) and fusion (content is drawn from two or more source sentences). Humans often grab content from 1 or 2 document sentences when writing a summary sentence.}
\label{fig:fuse}
\vspace{-0.2in}
\end{figure}

Abstractive summarization aims at presenting the main points of an article in a succinct and coherent manner. 
To achieve this goal, a proficient editor can rewrite a source sentence into a more succinct form by dropping inessential sentence elements such as prepositional phrases and adjectives.
She can also choose to fuse multiple source sentences into one by reorganizing the points in a coherent manner.
In fact, it appears to be common practice to summarize by either compressing single sentences or fusing multiple sentences.
We investigate this hypothesis by analyzing human-written abstracts contained in three large datasets: DUC-04~\cite{Over:2004}, CNN/Daily Mail~\cite{Hermann:2015}, and XSum~\cite{Narayan:2018:EMNLP}.
For every summary sentence, we find its \emph{ground-truth set} containing one or more source sentences that exhibit a high degree of similarity with the summary sentence (details in \S\ref{sec:data}). 
As shown in Figure~\ref{fig:fuse}, across the three datasets, 60-85\% of summary sentences are generated by fusing one or two source sentences.

Selecting summary-worthy sentences has been studied in the literature, but there lacks a mechanism to weigh sentence singletons and pairs in a unified space.
Extractive methods focus on selecting sentence singletons using greedy~\cite{Carbonell:1998}, optimization-based~\cite{Gillick:2009:NAACL,Kulesza:2011,Cho:2019}, and (non-)autoregressive methods~\cite{Cheng:2016,Kedzie:2018}.
In contrast, existing sentence fusion studies tend to assume ground sets of source sentences are already provided, and the system fuses each set of sentences into a single one~\cite{Daume:2004,Filippova:2010,Thadani:2013:IJCNLP}.
There is thus a crucial gap between sentence selection and fusion to support summarizing by both compressing single sentences and fusing pairs. 
This paper attempts to bridge the gap by ranking singletons and pairs together by their likelihoods of producing summary sentences.

The selection of sentence singletons and pairs can bring benefit to neural abstractive summarization, as a number of studies seek to separate content selection from summary generation~\cite{Chen:2018:ACL,Hsu:2018,Gehrmann:2018,Lebanoff:2018}.
Content selection draws on domain knowledge to identify relevant content, while summary generation weaves together selected source and vocabulary words to form a coherent summary.
Despite having local coherence, system summaries can sometimes contain erroneous details~\cite{See:2017} and forged content~\cite{Cao:2018,Song:2018}. 
Separating the two tasks of content selection and summary generation allows us to closely examine the compressing and fusing mechanisms of an abstractive summarizer.

In this paper we propose a method to learn to select sentence singletons and pairs, which then serve as the basis for an abstractive summarizer to compose a summary sentence-by-sentence, where singletons are shortened (i.e., compressed) and pairs are merged (i.e., fused).
We exploit state-of-the-art neural representations and traditional vector space models to characterize singletons and pairs; we then provide suggestions on the types of representations useful for summarization.
Experiments are performed on both single- and multi-document summarization datasets, where we demonstrate the efficacy of selecting sentence singletons and pairs as well as its utility to abstractive summarization.
Our research contributions can be summarized as follows:

\begin{itemize}[topsep=5pt,itemsep=0pt,leftmargin=*]

\item the present study fills an important gap by selecting sentence singletons and pairs jointly, assuming a summary sentence can be created by either shortening a singleton or merging a pair.
Compared to abstractive summarizers that perform content selection implicitly, our method is flexible and can be extended to multi-document summarization where training data is limited;

\item we investigate the factors involved in representing sentence singletons and pairs.
We perform extensive experiments and report findings on sentence selection and abstraction.\footnote{We make our code and models publicly available at {\scriptsize\url{https://github.com/ucfnlp/summarization-sing-pair-mix}}}

\end{itemize}

\begin{table*}[t]
\setlength{\tabcolsep}{4pt}
\renewcommand{\arraystretch}{1.2}
\centering
\begin{scriptsize}
\begin{fontppl}
\begin{tabular}{|l|l|}
\hline
\textbf{Sentence Pair:} & \textbf{Merged Sentence:} \\
\textsf{(A)} The \textcolor{blue}{bombing killed 58} people. & \textcolor{red}{Pakistan denies its spy agency helped plan} \textcolor{blue}{bombing that}\\
\textsf{(B)} Wajid Shamsul Hasan, \textcolor{red}{Pakistan}'s high commissioner to Britain, and Hamid Gul, &
\textcolor{blue}{killed 58.} \\
former head of the ISI, firmly \textcolor{red}{denied the agency's involvement} in the attack.
& \\
\hline
\hline
\textbf{Sentence Singleton:} & \textbf{Compressed Sentence:}\\
\textsf{(A)} Pakistani \textcolor{green!55!black}{Maj. Gen. Athar Abbas said the report} ``unfounded and malicious'' and & \textcolor{green!55!black}{Maj. Gen. Athar Abbas said the report was an ``effort to}\\
\textcolor{green!55!black}{an ``effort to malign the ISI,''} -- Pakistan's directorate of inter-services intelligence. & \textcolor{green!55!black}{malign the ISI.''}\\
\hline
\end{tabular}
\end{fontppl}
\end{scriptsize}
\caption{Example sentence singleton and pair, before and after compression/merging.
}
\label{tab:singleton_pair}
\end{table*}

\section{Related Work}
\label{sec:related}

Content selection is integral to any summarization system.
Neural approaches to abstractive summarization often perform content selection jointly with surface realization using an encoder-decoder architecture~\cite{Rush:2015,Nallapati:2016,Chen:2016,Tan:2017,See:2017,Paulus:2017,Celikyilmaz:2018,Narayan:2018:EMNLP}.
Training these models end-to-end means learning to perform both tasks simultaneously and can require a massive amount of data that is unavailable and unaffordable for many summarization tasks.

Recent approaches emphasize the importance of separating content selection from summary generation for abstractive summarization. 
Studies exploit extractive methods to identify content words and sentences that should be part of the summary and use them to guide the generation of abstracts~\cite{Chen:2018:ACL,Gehrmann:2018,Lebanoff:2018}.  
On the other hand, surface lexical features have been shown to be effective in identifying pertinent content~\cite{Carenini:2006,Wong:2008,Galanis:2012}.
Examples include sentence length, position, centrality, word frequency, whether a sentence contains topic words, and others. 
The surface cues can also be customized for new domains relatively easily.
This paper represents a step forward in this direction, where we focus on developing lightweight models to select summary-worthy sentence singletons and pairs and use them as the basis for summary generation.

A succinct sentence can be generated by shortening or rewriting a lengthy source text.
Recent studies have leveraged neural encoder-decoder models to rewrite the first sentence of an article to a title-like summary~\cite{Nallapati:2016,Zhou:2017,Li:2017:DRGD,Song:2018,Guo:2018:ACL,Cao:2018:ACL}.
Compressive summaries can be generated in a similar vein by selecting important source sentences and then dropping inessential sentence elements such as prepositional phrases.
Before the era of deep neural networks it has been an active area of research, where sentence selection and compression can be accomplished using a pipeline or a joint model~\cite{Daume:2002,Zajic:2007,Gillick:2009:NAACL,Wang:2013,Li:2013:EMNLP,Li:2014:EMNLP,Filippova:2015}. 
A majority of these studies focus on selecting and compressing sentence \emph{singletons} only.

A sentence can also be generated through fusing multiple source sentences. 
However, many aspects of this approach are largely underinvestigated, such as determining the set of source sentences to be fused, handling its large cardinality, and identifying the sentence relationships for performing fusion.
Previous studies assume a set of similar source sentences can be gathered by clustering sentences or by comparing to a reference summary sentence~\cite{Barzilay:2005,Filippova:2010,Shen:2010,Chenal:2016,Liao:2018}; but these methods can be suboptimal.
Joint models for sentence selection and fusion implicitly perform content planning~\cite{Martins:2009,Kirkpatrick:2011,Bing:2015,Durrett:2016} and there is limited control over which sentences are merged and how.

In contrast, this work attempts to teach the system to determine if a sentence singleton or a pair should be selected to produce a summary sentence.
A sentence pair (\textsf{A}, \textsf{B}) is preferred over its consisting sentences if they carry complementary content. 
Table~\ref{tab:singleton_pair} shows an example.
Sentence \textsf{B} contains a reference (``\emph{the attack}'') and \textsf{A} contains a more complete description for it (``\emph{bombing that killed 58}'').  
Sentences \textsf{A} and \textsf{B} each contain certain valuable information, and an appropriate way to merge them exists.
As a result, a sentence pair can be scored higher than a singleton given the content it carries and compatibility of its consisting sentences. 
In the following we describe methods to represent singletons and pairs in a unified framework and scoring them for summarization.

\section{Our Model}
\label{sec:our_approach}

We present the first attempt to transform sentence singletons and pairs to real-valued vector representations capturing semantic salience so that they can be measured against each other (\S\ref{sec:selection}). 
This is a nontrivial task, as it requires a direct comparison of texts of varying length---a pair of sentences is almost certainly longer than a single sentence. 
For sentence pairs, the representations are expected to further encode sentential semantic compatibility.
In \S\ref{sec:summarization}, we describe our method to utilize highest scoring singletons and pairs to a neural abstractive summarizer to generate summaries.

\subsection{Scoring Sentence Singletons and Pairs}
\label{sec:selection}

Given a document or set of documents, we create a set $\mathcal{D}$ of singletons and pairs by gathering all single sentences and arbitrary pairs of them.
We refer to a singleton or pair in the set as an \emph{instance}. 
The sentences in a pair are arranged in order of their appearance in the document or by date of documents. 
Let \textsf{N} be the number of single sentences in the input document(s), a complete set of singletons and pairs will contain $|\mathcal{D}|$=$\textstyle\frac{\textsf{N}(\textsf{N}-1)}{2}$+\textsf{\small N} instances.
Our goal is to score each instance based on the amount of summary-worthy content it conveys.
Despite their length difference, a singleton can be scored higher than a pair if it contains a significant amount of salient content.
Conversely, a pair can outweigh a singleton if its component sentences are salient and compatible with each other. 

Building effective representations for singletons and pairs is therefore of utmost importance.
We attempt to build a vector representation for each instance.
The representation should be invariant to the instance type, i.e., a singleton or pair.
In this paper we exploit the BERT architecture~\cite{Devlin:2018} to learn instance representations.
The representations are fine-tuned for a classification task predicting whether a given instance contains content used in human-written summary sentences (details for ground-truth creation in \S\ref{sec:data}).

\vspace{0.05in}
\textbf{BERT}\quad
BERT supports our goal of encoding singletons and pairs indiscriminately. 
It introduces two pretraining tasks to build deep contextual representations for words and sequences. 
A sequence can be a single sentence (\textsf{A}) or pair of sentences (\textsf{A}+\textsf{B}).\footnote{In the original BERT paper~\cite{Devlin:2018}, a ``sentence'' is used in a general sense to denote an arbitrary span of contiguous text;
we refer to an actual linguistic sentence.
}
The first task predicts \emph{missing words} in the input sequence.
The second task predicts if \textsf{B} is the \emph{next sentence} following \textsf{A}. 
It requires the vector representation for (\textsf{A}+\textsf{B}) to capture the coherence of two sentences.
As coherent sentences can often be fused together, we conjecture that the second task is particularly suited for our goal. 

Concretely, BERT constructs an input sequence by prepending a singleton or pair with a ``\textsf{\footnotesize[CLS]}'' symbol and delimiting the two sentences of a pair with ``\textsf{\footnotesize[SEP]}.'' 
The representation learned for the \textsf{\footnotesize[CLS]} symbol is used as an aggregate sequence representation for the later classification task.
We show an example input sequence in Eq.~(\ref{eq:w_i}). In the case of a singleton,  $w_i^\textsf{B}$ are padding tokens.
{\medmuskip=1mu
\thinmuskip=1mu
\thickmuskip=1mu
\nulldelimiterspace=0pt
\scriptspace=0pt
\begin{align*}
& \{w_i\} = \scriptsize\textsf{[CLS]}\;, w_1^\textsf{A}, w_2^\textsf{A}, \ldots, \;\scriptsize\textsf{[SEP]}\;, w_1^\textsf{B}, w_2^\textsf{B}, \ldots, \;\scriptsize\textsf{[SEP]}
\numberthis\label{eq:w_i}\\
& \mathbf{e}_i = \mathbf{e}_{\mbox{\scriptsize w}}(w_i) + \mathbf{e}_{\mbox{\scriptsize sgmt}}(w_i) + \mathbf{e}_{\mbox{\scriptsize wpos}}(w_i) + \mathbf{e}_{\mbox{\scriptsize spos}}(w_i)
\numberthis\label{eq:e_i}
\end{align*}}
\vspace{-0.2in}

In Eq.~(\ref{eq:e_i}), each token $w_i$ is characterized by an input embedding $\mathbf{e}_i$, calculated as the element-wise sum of the following embeddings: 
\begin{itemize}[topsep=5pt,itemsep=0pt,leftmargin=*]

\item $\mathbf{e}_{\mbox{\scriptsize w}}(w_i)$ is a \emph{token} embedding;

\item $\mathbf{e}_{\mbox{\scriptsize sgmt}}(w_i)$ is a \emph{segment} embedding, signifying whether $w_i$ comes from sentence \textsf{A} or \textsf{B}.

\item $\mathbf{e}_{\mbox{\scriptsize wpos}}(w_i)$ is a \emph{word position} embedding indicating the index of $w_i$ in the input sequence;

\item we introduce $\mathbf{e}_{\mbox{\scriptsize spos}}(w_i)$ to be a \emph{sentence position} embedding;
if $w_i$ is from sentence \textsf{A} (or \textsf{B}), $\mathbf{e}_{\mbox{\scriptsize spos}}(w_i)$ is the embedding indicating the index of sentence \textsf{A} (or \textsf{B}) in the original document. 

\end{itemize}

Intuitively, these embeddings mean that, the extent to which a word contributes to the sequence (\textsf{A}+\textsf{B}) representation depends on these factors: (i) word salience, (ii) importance of sentences \textsf{A} and \textsf{B}, (iii) word position in the sequence, and, (iv) sentence position in the document. 
These factors coincide with heuristics used in summarization literature~\cite{Nenkova:2011}, where leading sentences of a document and the first few words of a sentence are more likely to be included in the summary. 

The input embeddings are then fed to a multi-layer and multi-head attention architecture to build deep contextual representations for tokens.
Each layer employs a Transformer block~\cite{Vaswani:2017}, which introduces a self-attention mechanism that allows each hidden state $\mathbf{h}_i^l$ to be compared with every other hidden state of the same layer $[\mathbf{h}_1^l, \mathbf{h}_2^l, \ldots, \mathbf{h}_\textsf{\scriptsize N}^l]$ using a parallelizable, multi-head attention mechanism (Eq.~(\ref{eq:h_1}-\ref{eq:h_l})).
\begin{align*}
\mathbf{h}_i^1 &= f_{\mbox{\scriptsize self-attn}}^{1}(\mathbf{e}_i, [\mathbf{e}_1, \mathbf{e}_2, \ldots, \mathbf{e}_\textsf{\scriptsize N}])
\numberthis\label{eq:h_1}\\
\mathbf{h}_i^{l+1} &= f_{\mbox{\scriptsize self-attn}}^{l+1}(\mathbf{h}_i^l, [\mathbf{h}_1^l, \mathbf{h}_2^l, \ldots, \mathbf{h}_\textsf{\scriptsize N}^l])
\numberthis\label{eq:h_l}
\end{align*}

The representation at final layer $\textsf{L}$ for the \textsf{\footnotesize[CLS]} symbol is used as the sequence representation  $\mathbf{h}_{\textsf{\scriptsize [CLS]}}^\textsf{L}$.
The representations can be fine-tuned with an additional output layer to generate state-of-the-art results on a wide range of tasks including reading comprehension and natural language inference.  
We use the pretrained BERT base model and fine-tune it on our specific task of predicting if an instance (a singleton or pair) $p_{\scriptsize\mbox{inst}} = \sigma(\mathbf{w}^\top \mathbf{h}_{\textsf{\scriptsize [CLS]}}^\textsf{L})$ 
is an appropriate one, i.e., belonging to the ground-truth set of summary instances for a given document.
At test time, the architecture indiscriminately encodes a mixed collection of sentence singletons/pairs. 
We then obtain a likelihood score for each instance.
This framework is thus a first effort to build semantic representations for singletons and pairs capturing informativeness and semantic compatibility of two sentences.

\vspace{0.05in}
\textbf{VSM}\quad
We are interested in contrasting BERT with the traditional vector space model~\cite{Manning:2008} for representing singletons and pairs. 
BERT learns instance representations by attending to important content words, 
where the importance is signaled by word and position embeddings as well as pairwise word relationships.
Nonetheless, it remains an open question whether BERT can successfully weave the meaning of \emph{topically important words} into representations. 
A word ``border'' is topically important if the input document discusses border security.
A topic word is likely to be repeatedly mentioned in the input document but less frequently elsewhere.
Because sentences containing topical words are often deemed summary-worthy~\cite{Hong:2014:EACL}, 
it is desirable to represent sentence singletons and pairs based on the amount of topical content they convey.

VSM represents each sentence as a sparse vector. 
Each dimension of the vector corresponds to an $n$-gram weighted by its TF-IDF score.
A high TF-IDF score suggests the $n$-gram is important to the topic of discussion.
We further strengthen the sentence vector with position and centrality information, 
i.e., the sentence position in the document and the cosine similarity between the sentence and document vector. 
We obtain a document vector by averaging over its sentence vectors,
and we similarly obtain a vector for a pair of sentences. 
We use VSM representations as a baseline to contrast its performance with distributed representations from BERT.
To score singletons and pairs, we use the LambdaMART model\footnote{\scriptsize\url{https://sourceforge.net/p/lemur/wiki/RankLib/}} which has demonstrated success on related NLP tasks~\cite{Chen:2016:CNN};
it also fits our requirements of ranking singletons and pairs indiscriminately.

\subsection{Generating Summaries}
\label{sec:summarization}

We proceed by performing a preliminary investigation of summary generation from singletons and pairs; they are collectively referred to as \emph{instances}.
In the previous section, a set of summary instances is selected from a document.
These instances are treated as ``raw materials'' for a summary; they are fed to a neural abstractive summarizer which processes them into summary sentences via fusion and compression.
This strategy allows us to separately evaluate the contributions from instance selection and summary composition.

We employ the MMR principle~\cite{Carbonell:1998} to select a set of highest scoring and non-redundant instances.
The method adds an instance $\hat{P}$ to the summary $\mathcal{S}$ iteratively per Eq.~(\ref{eq:mmr}) until a length threshold has been reached.
Each instance is weighted by a linear combination of its importance score $\mathcal{I}(P_k)$, obtained by BERT or VSM, and its redundancy score $\mathcal{R}(P_k)$, computed as the cosine similarity between the instance and partial summary.
$\lambda$ is a balancing factor between importance and redundancy.\footnote{We use a coefficient $\lambda$ of 0.6.}
Essentially, MMR prevents the system from selecting instances that are too similar to ones already selected.
\begin{align*}
\hat{P} = \argmax_{P_k \in \mathcal{D} \setminus \mathcal{S}} \Big[\lambda \mathcal{I}(P_k) - (1-\lambda) \mathcal{R}(P_k) \Big]
\numberthis\label{eq:mmr}
\end{align*}

Composing a summary from selected instances is a non-trivial task. 
As a preliminary investigation of summary composition, we make use of pointer-generator (PG) networks~\cite{See:2017} to compress/fuse sentences into summary sentences. PG is a sequence-to-sequence model that has achieved state-of-the-art performance in abstractive summarization by having the ability to both copy tokens from the document or generate new tokens from the vocabulary. When trained on document-summary pairs, the model has been shown to remove unnecessary content from sentences and can merge multiple sentences together.

\begin{figure}[t]
\centering
\includegraphics[width=2.8in]{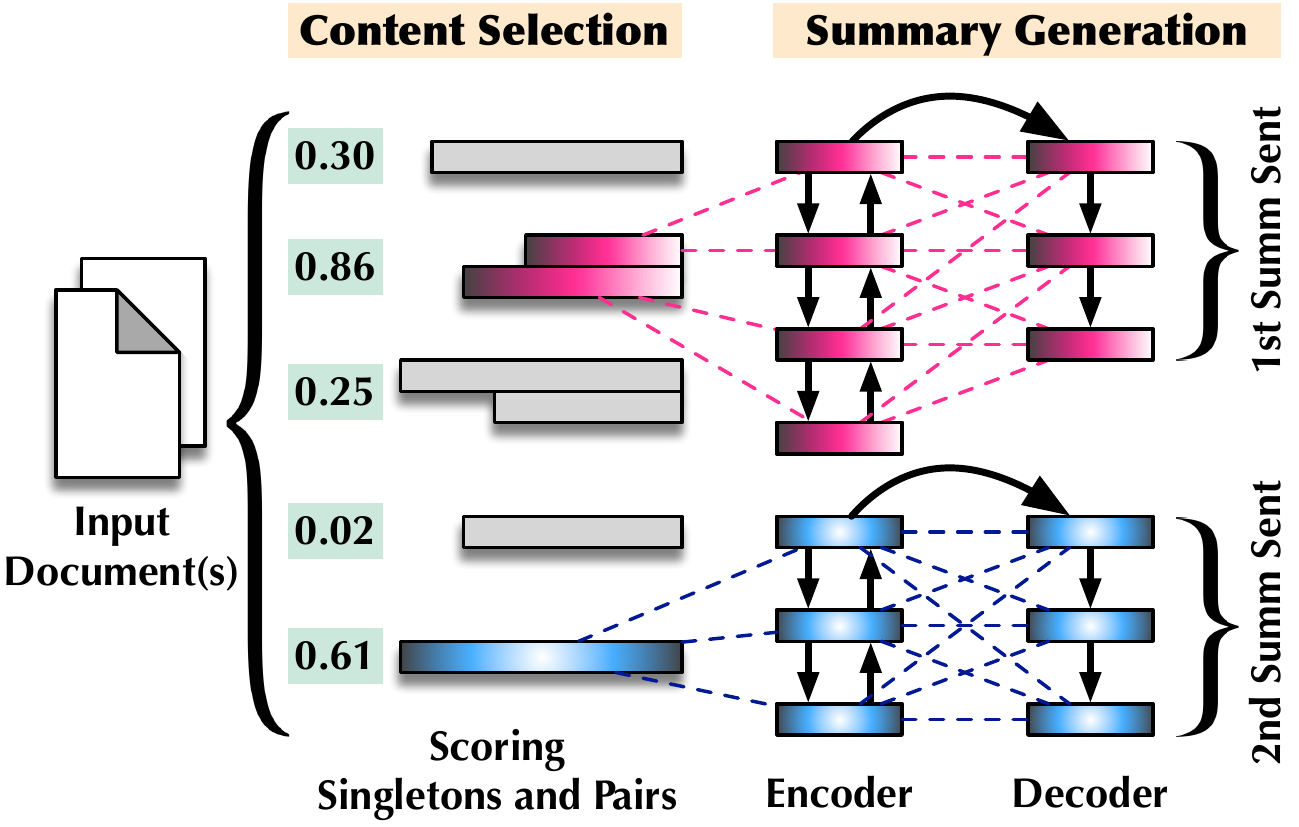}
\caption{System architecture. In this example, a sentence pair is chosen (red) and then merged to generate the first summary sentence. Next, a sentence singleton is selected (blue) and compressed for the second summary sentence.}
\label{fig:architecture}
\vspace{-0.18in}
\end{figure}

In this work, rather than training on document-summary pairs, we train PG exclusively on ground-truth instances. This removes most of the responsibility of content selection, and allows it to focus its efforts on merging the sentences. We use instances derived from human summaries (\S\ref{sec:data}) to train the network, which includes a sentence singleton or pair along with the ground-truth compressed/merged sentence.
At test time, the network receives an instance from BERT or VSM and outputs a summary sentence, then repeats this process to generate several sentences.
In Figure~\ref{fig:architecture} we present an illustration of the system architecture.

\section{Data}
\label{sec:data}

Our method does not require a massive amount of annotated data.
We thus report results on single- and multi-document summarization datasets.

We experiment with 
(i) {XSum}~\cite{Narayan:2018:EMNLP}, a new dataset created for extreme, abstractive summarization. 
The task is to reduce a news article to a short, one-sentence summary.
Both source articles and reference summaries are gathered from the BBC website.
The training set contains about 204k article-summary pairs and the test contains 11k pairs.
(ii) {CNN/DM}~\cite{Hermann:2015}, an abstractive summarization dataset frequently exploited by recent studies.
The task is to reduce a news article to a multi-sentence summary (4 sentences on average).
The training set contains about 287k article-summary pairs and the test set contains 11k pairs.
We use the non-anonymzied version of the dataset.
(iii) {DUC-04}~\cite{Over:2004}, a benchmark multi-document summarization dataset.
The task is to create an abstractive summary (5 sentences on average) from a set of 10 documents discussing a given topic.
The dataset contains 50 sets of documents used for testing purpose only.
Each document set is associated with four human reference summaries.

We build a training set for both tasks of content selection and summary generation.
This is done by creating ground-truth sets of instances based on document-summary pairs.
Each document and summary pair $(D,S)$ is a collection of sentences $D = \{d_1, d_2, ..., d_M\}$ and $S = \{s_1, s_2, ..., s_N\}$. We wish to associate each summary sentence $s_n$ with a subset of the document sentences  $\tilde{D} \subseteq D$, which are the sentences that are merged to form $s_n$. Our method chooses multiple sentences that work together to capture the most overlap with summary sentence $s_n$, in the following way. 

We use averaged ROUGE-1, -2, -L scores~\cite{Lin:2004} to represent sentence similarity. The source sentence most similar to $s_n$ is chosen, which we call $\tilde{d_1}$. All shared words are then removed from $s_n$ to create $s'_n$, effectively removing all information already captured by $\tilde{d_1}$. A second source sentence $\tilde{d_2}$ is selected that is most similar to the remaining summary sentence $s'_n$, and shared words are again removed from $s'_n$ to create $s''_n$. This process of sentence selection and overlap removal is repeated until no remaining sentences have at least two overlapping content words (words that are non-stopwords or punctuation) with $s_n$. The result is referred to as a ground-truth set $(s_n, \tilde{D})$ where $\tilde{D} = \{\tilde{d_1}, \tilde{d_2}, ..., \tilde{d}_{|\tilde{D}|}\}$. To train the models, $\tilde{D}$ is limited to one or two sentences because it captures the large majority of cases. All empty ground-truth sets are removed, and only the first two sentences are chosen for all ground-truth sets with more than two sentences.
A small number of summary sentences have empty ground-truth sets, corresponding to 2.85\%, 9.87\%, 5.61\% of summary sentences in CNN/DM, XSum, and DUC-04 datasets.
A detailed plot of the ground-truth set size is illustrated in Figure~\ref{fig:fuse}, and samples of the ground-truth are found in the supplementary.

We use the standard train/validation/test splits for both CNN/Daily Mail and XSum. We train our models on ground-truth sets of instances created from the training sets and tune hyperparameters using instances from the validation sets. DUC-04 is a test-only dataset, so we use the models trained on CNN/Daily Mail to evaluate DUC-04. Because the input is in the form of multiple documents, we select the first 20 sentences from each document and concatenate them together into a single mega-document \cite{Lebanoff:2018}. For the sentence position feature, we keep the sentence positions from the original documents. This handling of sentence position, along with other features that are invariant to the input type, allows us to effectively train on single-document inputs and transfer to the multi-document setting.

\begin{table*}
\setlength{\tabcolsep}{7pt}
\renewcommand{\arraystretch}{1.1}
\centering
\begin{small}
\begin{tabular}{|l|l||rrr|rrr||rrr|}
\hline
& & \multicolumn{3}{c|}{\textbf{Primary}} & \multicolumn{3}{c||}{\textbf{Secondary}} & \multicolumn{3}{c|}{\textbf{All}}\\
& \textbf{System} & \textbf{P} & \textbf{R} & \textbf{F} & \textbf{P} & \textbf{R} & \textbf{F} & \textbf{P} & \textbf{R} & \textbf{F}\\
\hline
\hline
\multirow{8}{*}{\rotatebox[origin=c]{90}{\textsf{CNN/Daily Mail}}} &
\textsc{Lead}-Baseline & 31.9 & 38.4 & 34.9 & 10.7 & 34.3 & 16.3 & 39.9 & 37.3 & 38.6\\
& SumBasic{\scriptsize~\cite{Vanderwende:2007}} & 15.2 & 17.3 & 16.2 & 5.3 & 15.8 & 8.0 & 19.6 & 16.9 & 18.1\\
& KL-Summ{\scriptsize~(Haghighi et al., 2009)\nocite{Haghighi:2009}} & 15.7 & 17.9 & 16.7 & 5.4 & 15.9 & 8.0 & 20.0 & 17.4 & 18.6\\
& LexRank{\scriptsize~\cite{Erkan:2004}} & 22.0 & 25.9 & 23.8 & 7.2 & 21.4 & 10.7 & 27.5 & 24.7 & 26.0\\
& VSM-SingOnly (This work) & 30.8 & 36.9 & 33.6 & 9.8 & 34.4 & 15.2 & 39.5 & 35.7 & 37.5\\
& VSM-SingPairMix (This work) & 27.0 & 46.5 & 34.2 & 9.0 & 42.1 & 14.9 & 34.0 & 45.4 & 38.9\\
& BERT-SingOnly (This work) & \textbf{35.3} & 41.9 & 38.3 & 9.8 & 32.5 & 15.1 & 44.0 & 38.6 & 41.1\\
& BERT-SingPairMix (This work) & 33.6 & \textbf{67.1} & \textbf{44.8} & \textbf{13.6} & \textbf{70.2} & \textbf{22.8} & \textbf{44.7} & \textbf{68.0} & \textbf{53.9}\\
\hline
\hline
\multirow{8}{*}{\rotatebox[origin=c]{90}{\textsf{XSum}}} &
\textsc{Lead}-Baseline & 8.5 & 9.4 & 8.9 & 5.3 & 9.5 & 6.8 & 13.8 & 9.4 & 11.2\\
& SumBasic{\scriptsize~\cite{Vanderwende:2007}} & 8.7 & 9.7 & 9.2 & 5.0 & 8.9 & 6.4 & 13.7 & 9.4 & 11.1\\
& KL-Summ{\scriptsize~(Haghighi et al., 2009)\nocite{Haghighi:2009}} & 9.2 & 10.2 & 9.7 & 5.0 & 8.9 & 6.4 & 14.2 & 9.7 & 11.5\\
& LexRank{\scriptsize~\cite{Erkan:2004}} & 9.7 & 10.8 & 10.2 & 5.5 & 9.8 & 7.0 & 15.2 & 10.4 & 12.4\\
& VSM-SingOnly (This work) & 12.3 & 14.1 & 13.1 & 3.8 & 11.0 & 5.6 & 17.9 & 12.0 & 14.4\\ 
& VSM-SingPairMix (This work) & 10.1 & 22.6 & 13.9 & 4.2 & 17.4 & 6.8 & 14.3 & 20.8 & 17.0\\ 
& BERT-SingOnly (This work) & 24.2 & 26.1 & 25.1 & 6.6 & 16.7 & 9.5 & 35.3 & 20.8 & 26.2\\
& BERT-SingPairMix (This work) & \textbf{33.2} & \textbf{56.0} & \textbf{41.7} & \textbf{24.1} & \textbf{65.5} & \textbf{35.2} & \textbf{57.3} & \textbf{59.6} & \textbf{58.5}\\
\hline
\hline
\multirow{7}{*}{\rotatebox[origin=c]{90}{\textsf{DUC-04}}} &
\textsc{Lead}-Baseline & 6.0 & 4.8 & 5.3 & 2.8 & 3.8 & 3.2 & 8.8 & 4.4 & 5.9\\
& SumBasic{\scriptsize~\cite{Vanderwende:2007}} & 4.2 & 3.2 & 3.6 & 3.0 & 3.8 & 3.3 & 7.2 & 3.4 & 4.6\\
& KL-Summ{\scriptsize~(Haghighi et al., 2009)\nocite{Haghighi:2009}} & 5.6 & 4.5 & 5.0 & 2.8 & 3.8 & 3.2 & 8.0 & 4.2 & 5.5\\
& LexRank{\scriptsize~\cite{Erkan:2004}} & 8.5 & 6.7 & 7.5 & \textbf{4.8} & 6.5 & 5.5 & 12.1 & 6.6 & 8.6\\
& VSM-SingOnly (This work) & \textbf{18.0} & \textbf{14.7} & \textbf{16.2} & 3.6 & 8.4 & 5.0 & \textbf{23.6} & \textbf{11.8} & \textbf{15.7}\\ 
& VSM-SingPairMix (This work) & 3.8 & 6.2 & 4.7 & 3.6 & 11.4 & 5.5 & 7.4 & 8.0 & 7.7\\ 
& BERT-SingOnly (This work) & 8.4 & 6.5 & 7.4 & 2.8 & 5.3 & 3.7 & 15.6 & 6.6 & 9.2\\
& BERT-SingPairMix (This work) & 4.8 & 9.1 & 6.3 & 4.2 & \textbf{14.2} & \textbf{6.5} & 9.0 & 10.9 & 9.9\\
\hline
\end{tabular}
\end{small}
\caption{
Instance selection results;
evaluated for primary, secondary, and all ground-truth sentences.
Our BERT-SingPairMix method achieves strong performance owing to its capability of building effective representations for both singletons and pairs.
}
\label{tab:results_sent_extr}
\vspace{-0.1in}
\end{table*}

\section{Results}
\label{sec:results}

\vspace{0.05in}
\textbf{Evaluation Setup}\quad
In this section we evaluate our proposed methods on identifying summary-worthy instances including singletons and pairs.
We compare this scheme with traditional methods extracting only singletons, then introduce novel evaluation strategies to compare results. 
We exploit several strong extractive baselines:
(i) \textit{SumBasic}~\cite{Vanderwende:2007} extracts sentences by assuming words occurring frequently in a document have higher chances of being included in the summary;
(ii) \textit{KL-Sum}~\cite{Haghighi:2009} greedily adds sentences to the summary to minimize KL divergence;
(iii) \textit{LexRank}~\cite{Erkan:2004} estimates sentence importance based on eigenvector centrality in a document graph representation.
Further, we include the \textsc{Lead} method that selects the first \textsf{N} sentences from each document.
We then require all systems to extract \textsf{N} instances, i.e., either singletons or pairs, from the input document(s).\footnote{
We use \textsf{N}=4/1/5 respectively for the CNN/DM, XSum, and DUC-04 datasets. \textsf{N} is selected as the average number of sentences in reference summaries.
}

We compare system-identified instances with ground-truth instances, and in particular, we compare against the primary, secondary, and full set of ground-truth sentences.
A \emph{primary} sentence is defined as a ground-truth singleton or a sentence in a ground-truth pair that has the highest similarity to the reference summary sentence;
the other sentence in the pair is considered \emph{secondary}, which provides complementary information to the primary sentence.
E.g., let $\mathcal{S}^*$=\{(1, 2), 5, (8, 4), 10\} be a ground-truth set of instances, where numbers are sentence indices and the first sentence of each pair is primary. 
Our ground-truth primary set thus contains \{1, 5, 8, 10\}; secondary set contains \{2, 4\}; and the full set of ground-truth sentences contains \{1, 2, 5, 8, 4, 10\}.
Assume $\mathcal{S}$=\{(1, 2), 3, (4, 10), 15\} are system-selected instances.
We uncollapse all pairs to obtain a set of single sentences $\mathcal{S}$=\{1, 2, 3, 4, 10, 15\}, 
then compare them against the primary, secondary, and full set of ground-truth sentences to calculate precision, recall, and F1-measure scores.
This evaluation scheme allows a fair comparison of a variety of systems for instance selection, and assess their performance on identifying primary and secondary sentences respectively for summary generation.

\vspace{0.05in}
\textbf{Extraction Results}\quad
In Table~\ref{tab:results_sent_extr} we present instance selection results for the {CNN/DM}, {XSum}, and {DUC-04} datasets.
Our method builds representations for instances using either BERT or VSM (\S\ref{sec:selection}).
To ensure a thorough comparison, we experiment with selecting a mixed set of singletons and pairs (``SingPairMix'') as well as selecting singletons only (``SingOnly'').
On the CNN/DM and XSum datasets, we observe that selecting a mixed set of singletons and pairs based on BERT representations (BERT+SingPairMix) demonstrates the most competitive results.
It outperforms a number of strong baselines when evaluated on a full set of ground-truth sentences.
The method also performs superiorly on identifying secondary sentences.
For example, it increases recall scores for identifying secondary sentences from 33.8\% to 69.8\% (CNN/DM) and from 16.7\% to 65.3\% (XSum).
Our method is able to achieve strong performance on instance selection owing to BERT's capability of building effective representations for both singletons and pairs.
It learns to identify salient source content based on token and position embeddings and it encodes sentential semantic compatibility using the pretraining task of predicting the next sentence; both are valuable additions to summary instance selection.

Further, we observe that identifying summary-worthy singletons and pairs from multi-document inputs (DUC-04) appears to be more challenging than that of single-document inputs (XSum and CNN/DM). 
This distinction is not surprising given that for multi-document inputs, the system has a large and diverse search space where candidate singletons and pairs are gathered from a set of documents written by different authors.\footnote{For the DUC-04 dataset, we select top \textsf{K} sentences from each document (\textsf{K}=5) and pool them as candidate singletons. Candidate pairs consist of arbitrary combinations of singletons. For all datasets we perform downsampling to balance the number of positive and negative singletons (or pairs).}
We find that the BERT model performs consistently on identifying secondary sentences, and VSM yields considerable performance gain on selecting primary sentences. 
Both BERT and VSM models are trained on the CNN/DM dataset and applied to DUC-04 as the latter data are only used for testing.
Our findings suggest that the TF-IDF features of the VSM model are effective for multi-document inputs, as important topic words are usually repeated across documents and TF-IDF scores can reflect topical importance of words. 
This analysis further reveals that extending BERT to incorporate topical salience of words can be a valuable line of research for future work.

\begin{table}[t]
\setlength{\tabcolsep}{5pt}
\renewcommand{\arraystretch}{1.1}
\centering
\begin{small}
\begin{tabular}{|l|rrr|}
\hline
& \multicolumn{3}{c|}{\textsf{CNN/Daily Mail}}\\
\textbf{System} & \textbf{R-1} & \,\,\,\textbf{R-2} & \textbf{R-L} \\
\hline
\hline
SumBasic{\scriptsize~\cite{Vanderwende:2007}} & 34.11 & 11.13 & 31.14\\
KLSumm{\scriptsize~(Haghighi et al., 2009)\nocite{Haghighi:2009}} & 29.92 & 10.50 & 27.37\\
LexRank{\scriptsize~\cite{Erkan:2004}} & 35.34 & 13.31 & 31.93\\
PointerGen+Cov{\scriptsize~\cite{See:2017}} & 39.53 & 17.28 & 36.38\\
BERT-Abs w/ SS (This Work) & 35.49 & 15.12 & 33.03\\
BERT-Abs w/ PG (This Work) & 37.15 & 15.22 & 34.60\\
BERT-Extr (This Work) & \textbf{41.13} & \textbf{18.68} & \textbf{37.75}\\
\hline
GT-SingPairMix (This Work) & 48.73 & 26.59 & 45.29\\
\hline
\hline
& \multicolumn{3}{c|}{\textsf{XSum}}\\
\textbf{System} & \textbf{R-1} & \,\,\,\textbf{R-2} & \textbf{R-L} \\
\hline
\hline
SumBasic{\scriptsize~\cite{Vanderwende:2007}} & 18.56 & 2.91 & 14.88\\
KLSumm{\scriptsize~(Haghighi et al., 2009)\nocite{Haghighi:2009}} & 16.73 & 2.83 & 13.53\\
LexRank{\scriptsize~\cite{Erkan:2004}} & 17.95 & 3.00 & 14.30\\
BERT-Abs w/ PG (This Work) & \textbf{25.08} & \textbf{6.48} & \textbf{19.75}\\ 
BERT-Extr (This Work) & 23.53 & 4.54 & 17.23\\
\hline
GT-SingPairMix (This Work) & 27.90 & 7.31 & 21.04\\
\hline
\hline
& \multicolumn{3}{c|}{\textsf{DUC-04}}\\
\textbf{System} & \textbf{R-1} & \,\,\,\textbf{R-2} & \textbf{R-SU4} \\
\hline
\hline
SumBasic{\scriptsize~\cite{Vanderwende:2007}} & 29.48 & 4.25 & 8.64\\
KLSumm{\scriptsize~(Haghighi et al., 2009)\nocite{Haghighi:2009}} & 31.04 & 6.03 & 10.23 \\
LexRank{\scriptsize~\cite{Erkan:2004}} & {34.44} & {7.11} & {11.19} \\
Extract+Rewrite{\scriptsize~\cite{Song:2018}} & 28.90 & 5.33 & 8.76 \\
Opinosis{\scriptsize~\cite{Ganesan:2010}} & 27.07 & 5.03 & 8.63 \\
BERT-Abs w/ PG (This Work) & 27.95 & 4.13 & 7.75\\
BERT-Extr (This Work) & 30.49 & 5.12 & 9.05\\
\hline
GT-SingPairMix (This Work) & 41.42 & 13.67 & 16.38\\
\hline
\end{tabular}
\end{small}
\caption{Summarization results on various datasets. 
Whether abstractive summaries (BERT-Abst) outperform its extractive variant (BERT-Extr) appears to be related to the amount of sentence pairs selected by BERT-SingPairMix.
Selecting more pairs than singletons seems to hurt the abstractor.
}
\label{tab:results_summ}
\vspace{-0.15in}
\end{table}

\vspace{0.05in}
\textbf{Summarization Results}\quad
We present summarization results in Table~\ref{tab:results_summ}, where we assess both extractive and abstractive summaries generated by BERT-SingPairMix. 
We omit VSM results as they are not as competitive as BERT on instance selection for the mixed set of singletons and pairs.  
The extractive summaries ``BERT-Extr'' are formed by concatenating selected singletons and pairs for each document,
whereas ``GT-SingPairMix'' concatenates \emph{ground-truth} singletons and pairs; 
it provides an upper bound for any system generating a set of singletons and pairs as the summary. 
To assure fair comparison, we limit all extractive summaries to contain up to 100 words (40 words for XSum) for ROUGE evaluation\footnote{w/ ROUGE options: {\fontppl\scriptsize -n 2 -m -2 4 -w 1.2 -c 95 -r 1000 -l 100}}, where R-1, R-2, R-L, and R-SU4 are variants used to measure the overlap of unigrams, bigrams, longest common subsequences, and skip bigrams (with a maximum distance of 4) between system and reference summaries~\cite{Lin:2004}. 
The abstractive summaries are generated from the same singletons and pairs used to form system extracts. 
``BERT-Abs-PG'' generates an abstract by iteratively encoding singletons or pairs and decoding summary sentences using pointer-generator networks (\S\ref{sec:summarization}).\footnote{We include an additional in-house system ``BERT-Abs-SS'' for CNN/DM  that takes the same input but generates summary sentences using a tree-based decoder.}

Our BERT summarization systems achieve results largely on par with those of prior work.
It is interesting to observe that the extractive variant (BERT-Extr) can outperform its abstractive counterparts on DUC-04 and CNN/DM datasets, and vice versa on XSum. 
A close examination of the results reveals that whether abstractive summaries outperform appears to be related to the amount of sentence pairs selected by ``BERT-SingPairMix.''
Selecting more pairs than singletons seems to hurt the abstractor.
For example, BERT selects 100\% and 76.90\% sentence pairs for DUC-04 and CNN/DM respectively, and 28.02\% for XSum.
These results suggest that existing abstractors using encoder-decoder models may need to improve on sentence fusion. 
These models are trained to generate fluent sentences more than preserving salient source content, leading to important content words being skipped in generating summary sentences.
Our work intends to separate the tasks of sentence selection and summary generation, thus holding promise for improving compression and merging in the future. 
We present example system summaries in the supplementary.

 \begin{figure}[t]
\centering
\includegraphics[width=3in]{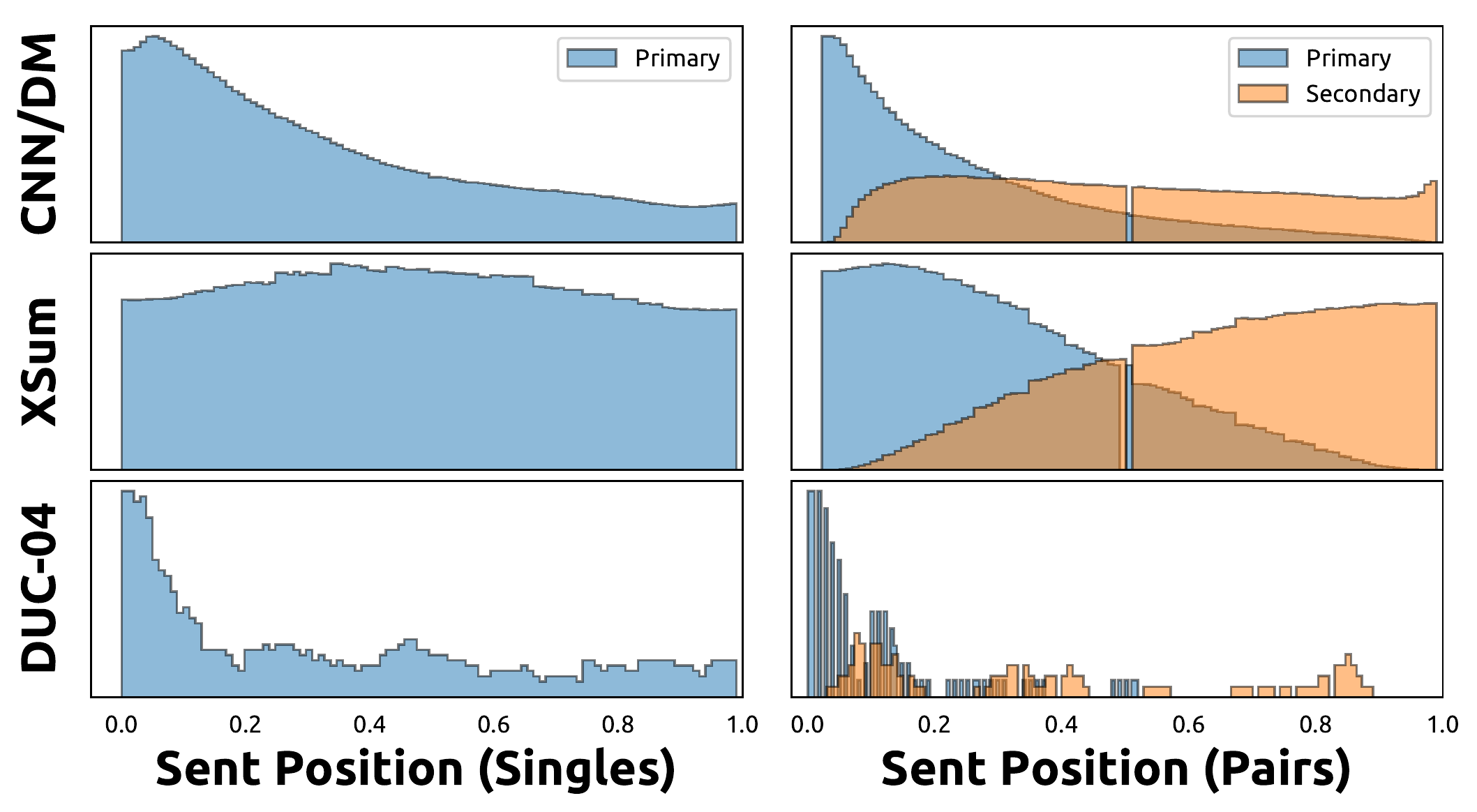}
\caption{Position of ground-truth singletons and pairs in a document.
The singletons of XSum can occur anywhere; the first and second sentence of a pair also appear far apart.
}
\label{fig:pos_doc}
\vspace{-0.15in}
\end{figure}

\begin{figure}[t]
\centering
\includegraphics[width=3in]{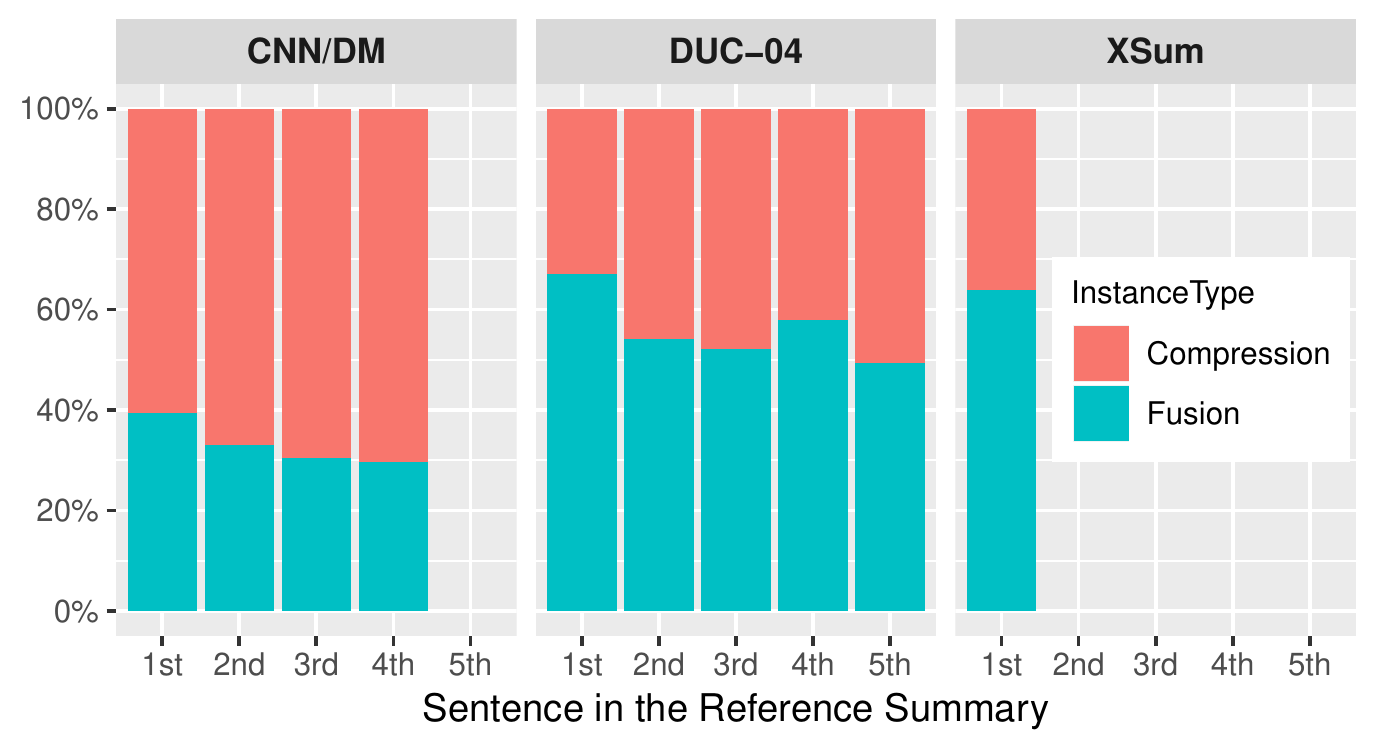}
\caption{
A sentence's \textit{position} in a human summary can affect whether or not it is created by compression or fusion.
}
\label{fig:pos_ref}
\vspace{-0.15in}
\end{figure}

\vspace{0.05in}
\textbf{Further analysis}\quad
In this section we perform a series of analyses to understand where summary-worthy content is located in a document and how humans order them into a summary.
Figure~\ref{fig:pos_doc} shows the position of ground-truth singletons and pairs in a document.
We observe that singletons of CNN/DM and DUC-04 tend to occur at the beginning of a document,
whereas singletons of XSum can occur anywhere.
We also find that the first and second sentence of a pair can appear far apart for XSum, but are closer for CNN/DM.
These findings suggest that selecting singletons and pairs for XSum can be more challenging than others, as indicated by the name ``extreme'' summarization.

Figure~\ref{fig:pos_ref} illustrates how humans choose to organize content into a summary.
Interestingly, we observe that a sentence's \textit{position} in a human summary affects whether or not it is created by compression or fusion.
The first sentence of a human-written summary is more likely than the following sentences to be a fusion of multiple source sentences. This is the case across all three datasets.
We conjecture that the first sentence of a summary is expected to give an overview of the document and needs to consolidate information from different parts. 
Other sentences of a human summary can be generated by simply shortening singletons. 
Our statistics reveal that DUC-04 and XSum summaries involve more fusion operations, exhibiting a higher level of abstraction than CNN/DM.

\section{Conclusion}
\label{sec:conclusion}

We present an investigation into the feasibility of scoring singletons and pairs according to their likelihoods of producing summary sentences.
Our framework is founded on the human process of selecting one or two sentences to merge together and it has the potential to bridge the gap between compression and fusion studies.
Our method provides a promising avenue for domain-specific summarization where content selection and summary generation are only loosely connected to reduce the costs of obtaining massive annotated data.

\section*{Acknowledgments}

We are grateful to the reviewers for their insightful comments that point to interesting future directions.
The authors also thank students in the UCF NLP group for useful discussions.

\bibliography{summ,abs_summ,fei}

\begin{thebibliography}{57}
\expandafter\ifx\csname natexlab\endcsname\relax\def\natexlab#1{#1}\fi

\bibitem[{Barzilay and McKeown(2005)}]{Barzilay:2005}
Regina Barzilay and Kathleen~R. McKeown. 2005.
\newblock \href
  {https://www.mitpressjournals.org/doi/pdf/10.1162/089120105774321091}
  {Sentence fusion for multidocument news summarization}.
\newblock \emph{Computational Linguistics}, 31(3).

\bibitem[{Berg-Kirkpatrick et~al.(2011)Berg-Kirkpatrick, Gillick, and
  Klein}]{Kirkpatrick:2011}
Taylor Berg-Kirkpatrick, Dan Gillick, and Dan Klein. 2011.
\newblock \href {https://dl.acm.org/citation.cfm?id=2002534} {Jointly learning
  to extract and compress}.
\newblock In \emph{Proceedings of the Annual Meeting of the Association for
  Computational Linguistics (ACL)}.

\bibitem[{Bing et~al.(2015)Bing, Li, Liao, Lam, Guo, and
  Passonneau}]{Bing:2015}
Lidong Bing, Piji Li, Yi~Liao, Wai Lam, Weiwei Guo, and Rebecca~J. Passonneau.
  2015.
\newblock \href {https://www.cs.cmu.edu/~lbing/pub/acl2015-bing.pdf}
  {Abstractive multi-document summarization via phrase selection and merging}.
\newblock In \emph{Proceedings of ACL}.

\bibitem[{Cao et~al.(2018{\natexlab{a}})Cao, Li, Li, and Wei}]{Cao:2018:ACL}
Ziqiang Cao, Wenjie Li, Sujian Li, and Furu Wei. 2018{\natexlab{a}}.
\newblock \href {https://aclweb.org/anthology/P18-1015} {Retrieve, rerank and
  rewrite: {S}oft template based neural summarization}.
\newblock In \emph{Proceedings of the Annual Meeting of the Association for
  Computational Linguistics (ACL)}.

\bibitem[{Cao et~al.(2018{\natexlab{b}})Cao, Wei, Li, and Li}]{Cao:2018}
Ziqiang Cao, Furu Wei, Wenjie Li, and Sujian Li. 2018{\natexlab{b}}.
\newblock \href {https://arxiv.org/abs/1711.04434} {Faithful to the original:
  {F}act aware neural abstractive summarization}.
\newblock In \emph{Proceedings of the AAAI Conference on Artificial
  Intelligence (AAAI)}.

\bibitem[{Carbonell and Goldstein(1998)}]{Carbonell:1998}
Jaime Carbonell and Jade Goldstein. 1998.
\newblock \href {https://dl.acm.org/citation.cfm?id=291025} {The use of {MMR},
  diversity-based reranking for reordering documents and producing summaries}.
\newblock In \emph{Proceedings of the International ACM SIGIR Conference on
  Research and Development in Information Retrieval (SIGIR)}.

\bibitem[{Carenini et~al.(2006)Carenini, Ng, and Pauls}]{Carenini:2006}
Giuseppe Carenini, Raymond Ng, and Adam Pauls. 2006.
\newblock \href {https://www.aclweb.org/anthology/E06-1039} {Multi-document
  summarization of evaluative text}.
\newblock In \emph{Proceedings of 11th Conference of the European Chapter of
  the Association for Computational Linguistics (EACL)}.

\bibitem[{Celikyilmaz et~al.(2018)Celikyilmaz, Bosselut, He, and
  Choi}]{Celikyilmaz:2018}
Asli Celikyilmaz, Antoine Bosselut, Xiaodong He, and Yejin Choi. 2018.
\newblock \href {https://arxiv.org/pdf/1803.10357.pdf} {Deep communicating
  agents for abstractive summarization}.
\newblock In \emph{Proceedings of the North American Chapter of the Association
  for Computational Linguistics (NAACL)}.

\bibitem[{Chen et~al.(2016{\natexlab{a}})Chen, Bolton, and
  Manning}]{Chen:2016:CNN}
Danqi Chen, Jason Bolton, and Christopher~D. Manning. 2016{\natexlab{a}}.
\newblock \href {https://www.aclweb.org/anthology/P16-1223} {A thorough
  examination of the cnn/daily mail reading comprehension task}.
\newblock In \emph{Proceedings of the 54th Annual Meeting of the Association
  for Computational Linguistics (ACL)}.

\bibitem[{Chen et~al.(2016{\natexlab{b}})Chen, Zhu, Ling, Wei, and
  Jiang}]{Chen:2016}
Qian Chen, Xiaodan Zhu, Zhen-Hua Ling, Si~Wei, and Hui Jiang.
  2016{\natexlab{b}}.
\newblock \href {https://arxiv.org/abs/1610.08462} {Distraction-based neural
  networks for document summarization}.
\newblock In \emph{Proceedings of the Twenty-Fifth International Joint
  Conference on Artificial Intelligence (IJCAI)}.

\bibitem[{Chen and Bansal(2018)}]{Chen:2018:ACL}
Yen-Chun Chen and Mohit Bansal. 2018.
\newblock \href {https://arxiv.org/abs/1805.11080} {Fast abstractive
  summarization with reinforce-selected sentence rewriting}.
\newblock In \emph{Proceedings of the Annual Meeting of the Association for
  Computational Linguistics (ACL)}.

\bibitem[{Chenal and Cheung(2016)}]{Chenal:2016}
Victor Chenal and Jackie Chi~Kit Cheung. 2016.
\newblock \href {https://www.aclweb.org/anthology/C16-1101} {Predicting
  sentential semantic compatibility for aggregation in text-to-text
  generation}.
\newblock In \emph{Proceedings of the International Conference on Computational
  Linguistics (COLING)}.

\bibitem[{Cheng and Lapata(2016)}]{Cheng:2016}
Jianpeng Cheng and Mirella Lapata. 2016.
\newblock \href {https://www.aclweb.org/anthology/P16-1046} {Neural
  summarization by extracting sentences and words}.
\newblock In \emph{Proceedings of ACL}.

\bibitem[{Cho et~al.(2019)Cho, Lebanoff, Foroosh, and Liu}]{Cho:2019}
Sangwoo Cho, Logan Lebanoff, Hassan Foroosh, and Fei Liu. 2019.
\newblock Improving the similarity measure of determinantal point processes for
  extractive multi-document summarization.
\newblock In \emph{Proceedings of the Annual Meeting of the Association for
  Computational Linguistics (ACL)}.

\bibitem[{{Daum\'{e} III} and Marcu(2002)}]{Daume:2002}
Hal {Daum\'{e} III} and Daniel Marcu. 2002.
\newblock \href {https://www.aclweb.org/anthology/P02-1057} {A noisy-channel
  model for document compression}.
\newblock In \emph{Proceedings of the Annual Meeting of the Association for
  Computational Linguistics (ACL)}.

\bibitem[{{Daum\'{e} III} and Marcu(2004)}]{Daume:2004}
Hal {Daum\'{e} III} and Daniel Marcu. 2004.
\newblock \href {https://www.aclweb.org/anthology/W04-1016} {Generic sentence
  fusion is an ill-defined summarization task}.
\newblock In \emph{Proceedings of ACL Workshop on Text Summarization Branches
  Out}.

\bibitem[{Devlin et~al.(2018)Devlin, Chang, Lee, and Toutanova}]{Devlin:2018}
Jacob Devlin, Ming-Wei Chang, Kenton Lee, and Kristina Toutanova. 2018.
\newblock \href {https://arxiv.org/abs/1810.04805} {{BERT:} pre-training of
  deep bidirectional transformers for language understanding}.
\newblock \emph{arXiv:1810.04805}.

\bibitem[{Durrett et~al.(2016)Durrett, Berg-Kirkpatrick, and
  Klein}]{Durrett:2016}
Greg Durrett, Taylor Berg-Kirkpatrick, and Dan Klein. 2016.
\newblock \href {https://arxiv.org/abs/1603.08887} {Learning-based
  single-document summarization with compression and anaphoricity constraints}.
\newblock In \emph{Proceedings of the Association for Computational Linguistics
  (ACL)}.

\bibitem[{Erkan and Radev(2004)}]{Erkan:2004}
G\"{u}nes Erkan and Dragomir~R. Radev. 2004.
\newblock \href {https://www.aaai.org/Papers/JAIR/Vol22/JAIR-2214.pdf}
  {{LexRank}: {G}raph-based lexical centrality as salience in text
  summarization}.
\newblock \emph{Journal of Artificial Intelligence Research}.

\bibitem[{Filippova(2010)}]{Filippova:2010}
Katja Filippova. 2010.
\newblock \href {https://www.aclweb.org/anthology/C10-1037} {Multi-sentence
  compression: {F}inding shortest paths in word graphs}.
\newblock In \emph{Proceedings of the International Conference on Computational
  Linguistics (COLING)}.

\bibitem[{Filippova et~al.(2015)Filippova, Alfonseca, Colmenares, Kaiser, and
  Vinyals}]{Filippova:2015}
Katja Filippova, Enrique Alfonseca, Carlos Colmenares, Lukasz Kaiser, and Oriol
  Vinyals. 2015.
\newblock \href
  {https://static.googleusercontent.com/media/research.google.com/en//pubs/archive/43852.pdf}
  {Sentence compression by deletion with lstms}.
\newblock In \emph{Proceedings of the Conference on Empirical Methods in
  Natural Language Processing (EMNLP)}.

\bibitem[{Galanis et~al.(2012)Galanis, Lampouras, and
  Androutsopoulos}]{Galanis:2012}
Dimitrios Galanis, Gerasimos Lampouras, and Ion Androutsopoulos. 2012.
\newblock \href {https://www.aclweb.org/anthology/C12-1056} {Extractive
  multi-document summarization with integer linear programming and support
  vector regression}.
\newblock In \emph{Proceedings of the International Conference on Computational
  Linguistics (COLING)}.

\bibitem[{Ganesan et~al.(2010)Ganesan, Zhai, and Han}]{Ganesan:2010}
Kavita Ganesan, ChengXiang Zhai, and Jiawei Han. 2010.
\newblock \href {https://www.aclweb.org/anthology/C10-1039} {Opinosis: {A}
  graph-based approach to abstractive summarization of highly redundant
  opinions}.
\newblock In \emph{Proceedings of the International Conference on Computational
  Linguistics (COLING)}.

\bibitem[{Gehrmann et~al.(2018)Gehrmann, Deng, and Rush}]{Gehrmann:2018}
Sebastian Gehrmann, Yuntian Deng, and Alexander~M. Rush. 2018.
\newblock \href {https://arxiv.org/abs/1808.10792} {Bottom-up abstractive
  summarization}.
\newblock In \emph{Proceedings of the Conference on Empirical Methods in
  Natural Language Processing (EMNLP)}.

\bibitem[{Gillick and Favre(2009)}]{Gillick:2009:NAACL}
Dan Gillick and Benoit Favre. 2009.
\newblock \href {https://dl.acm.org/citation.cfm?id=1611640} {A scalable global
  model for summarization}.
\newblock In \emph{Proceedings of the NAACL Workshop on Integer Linear
  Programming for Natural Langauge Processing}.

\bibitem[{Guo et~al.(2018)Guo, Pasunuru, and Bansal}]{Guo:2018:ACL}
Han Guo, Ramakanth Pasunuru, and Mohit Bansal. 2018.
\newblock \href {https://arxiv.org/abs/1805.11004} {Soft, layer-specific
  multi-task summarization with entailment and question generation}.
\newblock In \emph{Proceedings of the Annual Meeting of the Association for
  Computational Linguistics (ACL)}.

\bibitem[{Haghighi and Vanderwende(2009)}]{Haghighi:2009}
Aria Haghighi and Lucy Vanderwende. 2009.
\newblock \href {https://www.aclweb.org/anthology/N09-1041} {Exploring content
  models for multi-document summarization}.
\newblock In \emph{Proceedings of the North American Chapter of the Association
  for Computational Linguistics (NAACL)}.

\bibitem[{Hermann et~al.(2015)Hermann, Kocisky, Grefenstette, Espeholt, Kay,
  Suleyman, and Blunsom}]{Hermann:2015}
Karl~Moritz Hermann, Tomas Kocisky, Edward Grefenstette, Lasse Espeholt, Will
  Kay, Mustafa Suleyman, and Phil Blunsom. 2015.
\newblock \href {https://arxiv.org/abs/1506.03340} {Teaching machines to read
  and comprehend}.
\newblock In \emph{Proceedings of Neural Information Processing Systems
  (NIPS)}.

\bibitem[{Hong and Nenkova(2014)}]{Hong:2014:EACL}
Kai Hong and Ani Nenkova. 2014.
\newblock \href {https://www.aclweb.org/anthology/E14-1075} {Improving the
  estimation of word importance for news multi-document summarization}.
\newblock In \emph{Proceedings of the Conference of the European Chapter of the
  Association for Computational Linguistics (EACL)}.

\bibitem[{Hsu et~al.(2018)Hsu, Lin, Lee, Min, Tang, and Sun}]{Hsu:2018}
Wan-Ting Hsu, Chieh-Kai Lin, Ming-Ying Lee, Kerui Min, Jing Tang, and Min Sun.
  2018.
\newblock \href {https://arxiv.org/abs/1805.06266} {A unified model for
  extractive and abstractive summarization using inconsistency loss}.
\newblock In \emph{Proceedings of the Annual Meeting of the Association for
  Computational Linguistics (ACL)}.

\bibitem[{Kedzie et~al.(2018)Kedzie, McKeown, and III}]{Kedzie:2018}
Chris Kedzie, Kathleen McKeown, and Hal~Daume III. 2018.
\newblock \href {https://arxiv.org/abs/1810.12343} {Content selection in deep
  learning models of summarization}.
\newblock In \emph{Proceedings of the Conference on Empirical Methods in
  Natural Language Processing (EMNLP)}.

\bibitem[{Kulesza and Taskar(2011)}]{Kulesza:2011}
Alex Kulesza and Ben Taskar. 2011.
\newblock \href {https://dl.acm.org/citation.cfm?id=3020597} {Learning
  determinantal point processes}.
\newblock In \emph{Proceedings of the Conference on Uncertainty in Artificial
  Intelligence (UAI)}.

\bibitem[{Lebanoff et~al.(2018)Lebanoff, Song, and Liu}]{Lebanoff:2018}
Logan Lebanoff, Kaiqiang Song, and Fei Liu. 2018.
\newblock \href {https://aclweb.org/anthology/D18-1446} {Adapting the neural
  encoder-decoder framework from single to multi-document summarization}.
\newblock In \emph{Proceedings of the Conference on Empirical Methods in
  Natural Language Processing (EMNLP)}.

\bibitem[{Li et~al.(2013)Li, Liu, Weng, and Liu}]{Li:2013:EMNLP}
Chen Li, Fei Liu, Fuliang Weng, and Yang Liu. 2013.
\newblock \href {https://www.aclweb.org/anthology/D13-1047} {Document
  summarization via guided sentence compression}.
\newblock In \emph{Proceedings of the 2013 Conference on Empirical Methods in
  Natural Language Processing (EMNLP)}.

\bibitem[{Li et~al.(2014)Li, Liu, Liu, Zhao, and Weng}]{Li:2014:EMNLP}
Chen Li, Yang Liu, Fei Liu, Lin Zhao, and Fuliang Weng. 2014.
\newblock \href {https://www.aclweb.org/anthology/D14-1076} {Improving
  multi-document summarization by sentence compression based on expanded
  constituent parse tree}.
\newblock In \emph{Proceedings of the Conference on Empirical Methods on
  Natural Language Processing (EMNLP)}.

\bibitem[{Li et~al.(2017)Li, Lam, Bing, and Wang}]{Li:2017:DRGD}
Piji Li, Wai Lam, Lidong Bing, and Zihao Wang. 2017.
\newblock \href {https://www.aclweb.org/anthology/D17-1222} {Deep recurrent
  generative decoder for abstractive text summarization}.
\newblock In \emph{Proceedings of the Conference on Empirical Methods in
  Natural Language Processing (EMNLP)}.

\bibitem[{Liao et~al.(2018)Liao, Lebanoff, and Liu}]{Liao:2018}
Kexin Liao, Logan Lebanoff, and Fei Liu. 2018.
\newblock \href {https://arxiv.org/abs/1806.05655} {Abstract meaning
  representation for multi-document summarization}.
\newblock In \emph{Proceedings of the International Conference on Computational
  Linguistics (COLING)}.

\bibitem[{Lin(2004)}]{Lin:2004}
Chin-Yew Lin. 2004.
\newblock \href {https://www.aclweb.org/anthology/W04-1013} {{ROUGE}: a package
  for automatic evaluation of summaries}.
\newblock In \emph{Proceedings of ACL Workshop on Text Summarization Branches
  Out}.

\bibitem[{Manning et~al.(2008)Manning, Raghavan, and
  Sch{\"u}tze}]{Manning:2008}
Christopher~D. Manning, Prabhakar Raghavan, and Hinrich Sch{\"u}tze. 2008.
\newblock \emph{Introduction to Information Retrieval}.
\newblock Cambridge University Press.

\bibitem[{Martins and Smith(2009)}]{Martins:2009}
Andre F.~T. Martins and Noah~A. Smith. 2009.
\newblock \href {https://www.aclweb.org/anthology/W09-1801} {Summarization with
  a joint model for sentence extraction and compression}.
\newblock In \emph{Proceedings of the ACL Workshop on Integer Linear
  Programming for Natural Language Processing}.

\bibitem[{Nallapati et~al.(2016)Nallapati, Zhou, dos Santos, Gulcehre, and
  Xiang}]{Nallapati:2016}
Ramesh Nallapati, Bowen Zhou, Cicero dos Santos, Caglar Gulcehre, and Bing
  Xiang. 2016.
\newblock \href {https://arxiv.org/abs/1602.06023} {Abstractive text
  summarization using sequence-to-sequence rnns and beyond}.
\newblock In \emph{Proceedings of SIGNLL}.

\bibitem[{Narayan et~al.(2018)Narayan, Cohen, and Lapata}]{Narayan:2018:EMNLP}
Shashi Narayan, Shay~B. Cohen, and Mirella Lapata. 2018.
\newblock \href {https://arxiv.org/abs/1808.08745} {Don't give me the details,
  just the summary! {T}opic-aware convolutional neural networks for extreme
  summarization}.
\newblock In \emph{Proceedings of the Conference on Empirical Methods in
  Natural Language Processing (EMNLP)}.

\bibitem[{Nenkova and McKeown(2011)}]{Nenkova:2011}
Ani Nenkova and Kathleen McKeown. 2011.
\newblock \href {https://www.nowpublishers.com/article/Details/INR-015}
  {Automatic summarization}.
\newblock \emph{Foundations and Trends in Information Retrieval}.

\bibitem[{Over and Yen(2004)}]{Over:2004}
Paul Over and James Yen. 2004.
\newblock \href {https://duc.nist.gov/pubs/2004slides/duc2004.intro.pdf} {An
  introduction to {DUC}-2004}.
\newblock \emph{National Institute of Standards and Technology}.

\bibitem[{Paulus et~al.(2017)Paulus, Xiong, and Socher}]{Paulus:2017}
Romain Paulus, Caiming Xiong, and Richard Socher. 2017.
\newblock \href {https://arxiv.org/abs/1705.04304} {A deep reinforced model for
  abstractive summarization}.
\newblock In \emph{Proceedings of the Conference on Empirical Methods in
  Natural Language Processing (EMNLP)}.

\bibitem[{Rush et~al.(2015)Rush, Chopra, and Weston}]{Rush:2015}
Alexander~M. Rush, Sumit Chopra, and Jason Weston. 2015.
\newblock \href {https://www.aclweb.org/anthology/D15-1044} {A neural attention
  model for sentence summarization}.
\newblock In \emph{Proceedings of EMNLP}.

\bibitem[{See et~al.(2017)See, Liu, and Manning}]{See:2017}
Abigail See, Peter~J. Liu, and Christopher~D. Manning. 2017.
\newblock \href {https://arxiv.org/abs/1704.04368} {Get to the point:
  {S}ummarization with pointer-generator networks}.
\newblock In \emph{Proceedings of the Annual Meeting of the Association for
  Computational Linguistics (ACL)}.

\bibitem[{Shen and Li(2010)}]{Shen:2010}
Chao Shen and Tao Li. 2010.
\newblock \href {https://www.aclweb.org/anthology/C10-1111} {Multi-document
  summarization via the minimum dominating set}.
\newblock In \emph{Proceedings of the International Conference on Computational
  Linguistics (COLING)}.

\bibitem[{Song et~al.(2018)Song, Zhao, and Liu}]{Song:2018}
Kaiqiang Song, Lin Zhao, and Fei Liu. 2018.
\newblock \href {https://aclweb.org/anthology/C18-1146} {Structure-infused copy
  mechanisms for abstractive summarization}.
\newblock In \emph{Proceedings of the International Conference on Computational
  Linguistics (COLING)}.

\bibitem[{Tan et~al.(2017)Tan, Wan, and Xiao}]{Tan:2017}
Jiwei Tan, Xiaojun Wan, and Jianguo Xiao. 2017.
\newblock \href {https://www.aclweb.org/anthology/P17-1108} {Abstractive
  document summarization with a graph-based attentional neural model}.
\newblock In \emph{Proceedings of the Annual Meeting of the Association for
  Computational Linguistics (ACL)}.

\bibitem[{Thadani and McKeown(2013)}]{Thadani:2013:IJCNLP}
Kapil Thadani and Kathleen McKeown. 2013.
\newblock \href {https://www.aclweb.org/anthology/I13-1198} {Supervised
  sentence fusion with single-stage inference}.
\newblock In \emph{Proceedings of the International Joint Conference on Natural
  Language Processing (IJCNLP)}.

\bibitem[{Vanderwende et~al.(2007)Vanderwende, Suzuki, Brockett, and
  Nenkova}]{Vanderwende:2007}
Lucy Vanderwende, Hisami Suzuki, Chris Brockett, and Ani Nenkova. 2007.
\newblock \href {https://www.cis.upenn.edu/~nenkova/papers/ipm.pdf} {Beyond
  {SumBasic}: {T}ask-focused summarization with sentence simplification and
  lexical expansion}.
\newblock \emph{Information Processing and Management}, 43(6):1606--1618.

\bibitem[{Vaswani et~al.(2017)Vaswani, Shazeer, Parmar, Uszkoreit, Jones,
  Gomez, Kaiser, and Polosukhin}]{Vaswani:2017}
Ashish Vaswani, Noam Shazeer, Niki Parmar, Jakob Uszkoreit, Llion Jones,
  Aidan~N. Gomez, Lukasz Kaiser, and Illia Polosukhin. 2017.
\newblock \href {https://arxiv.org/abs/1706.03762}
  {https://arxiv.org/abs/1706.03762}.
\newblock In \emph{Proceedings of the 31st Conference on Neural Information
  Processing Systems (NIPS)}.

\bibitem[{Wang et~al.(2013)Wang, Raghavan, Castelli, Florian, and
  Cardie}]{Wang:2013}
Lu~Wang, Hema Raghavan, Vittorio Castelli, Radu Florian, and Claire Cardie.
  2013.
\newblock \href {https://arxiv.org/abs/1606.07548} {A sentence compression
  based framework to query-focused multi-document summarization}.
\newblock In \emph{Proceedings of ACL}.

\bibitem[{Wong et~al.(2008)Wong, Wu, and Li}]{Wong:2008}
Kam-Fai Wong, Mingli Wu, and Wenjie Li. 2008.
\newblock \href {https://www.aclweb.org/anthology/C08-1124} {Extractive
  summarization using supervised and semi-supervised learning}.
\newblock In \emph{Proceedings of the International Conference on Computational
  Linguistics (COLING)}.

\bibitem[{Zajic et~al.(2007)Zajic, Dorr, Lin, and Schwartz}]{Zajic:2007}
David Zajic, Bonnie~J. Dorr, Jimmy Lin, and Richard Schwartz. 2007.
\newblock \href
  {http://users.umiacs.umd.edu/~jimmylin/publications/Zajic_etal_IPM2007.pdf}
  {Multi-candidate reduction: {S}entence compression as a tool for document
  summarization tasks}.
\newblock \emph{Information Processing and Management}.

\bibitem[{Zhou et~al.(2017)Zhou, Yang, Wei, and Zhou}]{Zhou:2017}
Qingyu Zhou, Nan Yang, Furu Wei, and Ming Zhou. 2017.
\newblock \href {https://arxiv.org/abs/1704.07073} {Selective encoding for
  abstractive sentence summarization}.
\newblock In \emph{Proceedings of the Annual Meeting of the Association for
  Computational Linguistics (ACL)}.

\end{thebibliography}
\bibliographystyle{acl_natbib}

\appendix
\section{Ground-truth Sets of Instances}

We performed a manual inspection over a subset of our ground-truth sets of singletons and pairs. 
Each sentence from a human-written summary is matched with one or two source sentences based on average ROUGE similarity 
(details in Section 4 of the paper). 
Tables \ref{tab:cnndm_labels}, \ref{tab:xsum_labels}, and \ref{tab:duc_labels} present randomly selected examples from CNN/Daily Mail, XSum, and DUC-04, respectively. Colored text represents overlapping tokens between sentences. Darker colors represent content from primary sentences, while lighter colors represent content from secondary sentences. Best viewed in color.

\begin{table*}[t]
\setlength{\tabcolsep}{5pt}
\renewcommand{\arraystretch}{1.1}
\centering
\begin{scriptsize}
\begin{fontppl}
\begin{tabular}{|p{3.7in}|p{2.2in}|}
\hline

\textbf{Selected Source Sentence(s)} & \textbf{Human Summary Sentence}
\\
\hline
\hline
an \hlc[1d]{inmate housed on the `` forgotten floor , ''} where many \hlc[1d]{mentally ill inmates are housed in miami} before trial .
&
\hlc[1d]{mentally ill inmates in miami are housed on the `` forgotten floor '' }
\\
\hline

\hlc[2d]{most} often , they face drug charges or charges of assaulting an officer -- charges that \hlc[2d]{judge steven leifman says are} usually \hlc[2d]{`` avoidable felonies . ''} 
&
\hlc[2d]{judge steven leifman says most are} there as a result \hlc[2d]{of `` avoidable felonies ''} 
\\
\hline
\hlc[3d]{`` i am the son of the president} . 

miami , florida -lrb- \hlc[3l]{cnn} -rrb- -- the ninth floor of the miami-dade pretrial detention \hlc[3l]{facility} is dubbed the `` forgotten floor . ''
&
while \hlc[3l]{cnn} tours \hlc[3l]{facility ,} patient shouts : \hlc[3d]{`` i am the son of the president} ''
\\
\hline
it 's brutally \hlc[4d]{unjust} , in his mind , \hlc[4d]{and he has} become a strong advocate \hlc[4d]{for changing} things in miami .

so , he \hlc[4l]{says} , the sheer volume is overwhelming \hlc[4l]{the system} , and the result is what we see on the ninth floor . 
&
leifman \hlc[4l]{says the system is} \hlc[4d]{unjust and he 's} fighting \hlc[4d]{for change .}
\\
\hline
\end{tabular}
\vspace{0.2in}

\begin{tabular}{|p{3.7in}|p{2.2in}|}
\hline

\textbf{Selected Source Sentence(s)} & \textbf{Human Summary Sentence}
\\
\hline
\hline
the average surface temperature \hlc[1d]{has warmed one degree} fahrenheit -lrb- 0.6 \hlc[1d]{degrees} celsius -rrb- during the last century , according to the national research council .
&
earth \hlc[1d]{has warmed one degree} in past 100 years .
\\
\hline
the reason most cited -- \hlc[2d]{by scientists} and scientific organizations -- for the current warming trend is an increase in the concentrations \hlc[2d]{of greenhouse gases} , which are in the atmosphere naturally and help keep the planet 's \hlc[2d]{temperature} at a comfortable level .

in the worst-case scenario , experts \hlc[2l]{say} oceans could \hlc[2l]{rise to} overwhelming and catastrophic levels , flooding cities and altering seashores . 
&
majority \hlc[2d]{of scientists} \hlc[2l]{say} \hlc[2d]{greenhouse gases are} causing \hlc[2d]{temperatures} \hlc[2l]{to rise} .
\\
\hline
a change in the earth 's orbit or the intensity of the sun 's radiation could change , triggering \hlc[3d]{warming or cooling .}

other scientists and observers , a minority compared to those who believe the warming trend is something ominous , \hlc[3l]{say} it is simply the latest shift in the cyclical patterns of a \hlc[3l]{planet} 's life . 
&
some critics \hlc[3l]{say planets} often in periods \hlc[3d]{of warming or cooling .} 
\\
\hline
\end{tabular}

\end{fontppl}
\end{scriptsize}
\caption{Sample of our ground-truth labels for singleton/pair instances from CNN/Daily Mail. Large chunks of text are copied straight out of the source sentences.
}
\label{tab:cnndm_labels}
\vspace{0in}
\end{table*}

\begin{table*}[t]
\setlength{\tabcolsep}{5pt}
\renewcommand{\arraystretch}{1.1}
\centering
\begin{scriptsize}
\begin{fontppl}

\begin{tabular}{|p{3.7in}|p{2.2in}|}
\hline

\textbf{Selected Source Sentence(s)} & \textbf{Human Summary Sentence}
\\
\hline
\hline
the premises , used by \hlc[1d]{east belfast} mp naomi long , \hlc[1d]{have been} targeted a number of times . 

army explosives experts were called out to deal with a suspect \hlc[1l]{package} at the \hlc[1l]{offices} on the newtownards road on friday night . 
&
a suspicious \hlc[1l]{package} left outside an alliance party \hlc[1l]{office} in \hlc[1d]{east belfast has been} declared a hoax .
\\
\hline
\end{tabular}
\vspace{0.1in}

\begin{tabular}{|p{3.7in}|p{2.2in}|}
\hline

\textbf{Selected Source Sentence(s)} & \textbf{Human Summary Sentence}
\\
\hline
\hline
nev edwards scored an early \hlc[1d]{try} for \hlc[1d]{sale} , before \hlc[1d]{castres} ' florian vialelle went \hlc[1d]{over} , but julien dumora 's \hlc[1d]{penalty} put the hosts 10-7 ahead at the break . 
&
a late \hlc[1d]{penalty try} gave \hlc[1d]{sale} victory \hlc[1d]{over castres at} stade pierre-antoine in their european challenge cup clash . 
\\
\hline
\end{tabular}
\vspace{0.1in}

\begin{tabular}{|p{3.7in}|p{2.2in}|}
\hline

\textbf{Selected Source Sentence(s)} & \textbf{Human Summary Sentence}
\\
\hline
\hline
speaking \hlc[1d]{in the} dáil , sinn féin leader gerry adams also \hlc[1d]{called} for \hlc[1d]{a commission of investigation} and said his party had `` little confidence \hlc[1d]{the government} is protecting the public interest '' . 

last year , \hlc[1l]{nama} sold its entire 850-property \hlc[1l]{loan portfolio} in \hlc[1l]{northern ireland to} the new york investment firm cerberus for more than \# 1bn . 
&
the irish \hlc[1d]{government has} rejected \hlc[1d]{calls} to set up \hlc[1d]{a commission of investigation} into the sale of \hlc[1l]{nama} 's \hlc[1l]{portfolio} of \hlc[1l]{loans in northern ireland} . 
\\
\hline
\end{tabular}

\end{fontppl}
\end{scriptsize}
\caption{Sample of our ground-truth labels for singleton/pair instances from XSum. Each article has only one summary sentences, and thus only one singleton or pair matched with it.
}
\label{tab:xsum_labels}
\vspace{-0.1in}
\end{table*}

\begin{table*}[t]
\setlength{\tabcolsep}{5pt}
\renewcommand{\arraystretch}{1.1}
\centering
\begin{scriptsize}
\begin{fontppl}

\begin{tabular}{|p{3.7in}|p{2.2in}|}
\hline

\textbf{Selected Source Sentence(s)} & \textbf{Human Summary Sentence}
\\
\hline
\hline
\hlc[1d]{hun sen} 's \hlc[1d]{cambodian} people 's \hlc[1d]{party} won 64 \hlc[1d]{of the} 122 parliamentary seats in july 's \hlc[1d]{elections ,} short \hlc[1d]{of the} two-thirds \hlc[1d]{majority} needed \hlc[1d]{to form a government} on \hlc[1d]{its own .} 
&
\hlc[1d]{cambodian elections ,} fraudulent according to opposition \hlc[1d]{parties ,} gave the cpp \hlc[1d]{of hun sen a} scant \hlc[1d]{majority} but not enough \hlc[1d]{to form its own government .} 
\\
\hline
\hlc[2d]{opposition leaders} prince norodom ranariddh and sam rainsy , citing hun sen 's threats to \hlc[2d]{arrest opposition} figures after two alleged attempts on his life , said they could not negotiate freely in cambodia and called \hlc[2d]{for talks} at sihanouk 's residence in beijing . 

cambodian leader hun sen has guaranteed the safety and political freedom of all politicians , trying to ease \hlc[2l]{the fears} of his rivals that they will be arrested or killed if they return to \hlc[2l]{the country} . 
&
\hlc[2d]{opposition leaders} \hlc[2l]{fearing} \hlc[2d]{arrest ,} or worse , fled and asked \hlc[2d]{for talks} outside \hlc[2l]{the country} . 
\\
\hline
the cambodian people 's party criticized a non-binding resolution passed earlier this month \hlc[4d]{by the} u.s. \hlc[4d]{house of} representatives \hlc[4d]{calling for an investigation} into \hlc[4d]{violations of} international humanitarian law allegedly committed \hlc[4d]{by hun sen .} 
&
the un found evidence of rights \hlc[4d]{violations by hun sen} prompting the us \hlc[4d]{house} to \hlc[4d]{call for an investigation .} 
\\
\hline
cambodian politicians expressed hope monday that a new partnership between the parties of strongman hun \hlc[5d]{sen and his rival , prince norodom ranariddh ,} in a coalition government would not \hlc[5d]{end} in more violence . 
&
the three-month governmental deadlock \hlc[5d]{ended} with han \hlc[5d]{sen and his} chief \hlc[5d]{rival , prince norodom ranariddh} sharing power . 
\\
\hline
citing hun \hlc[6d]{sen} 's threats to arrest opposition politicians following two alleged attempts on his life , ranariddh and \hlc[6d]{sam rainsy} have said they do not feel \hlc[6d]{safe} negotiating inside the country and asked the king to chair the summit at gis residence in beijing .

after a meeting between hun sen and the new french ambassador to \hlc[6l]{cambodia} , hun sen aide prak sokhonn said the cambodian leader had repeated calls for the opposition to \hlc[6l]{return} , but expressed concern that the international community may be asked for security \hlc[6l]{guarantees} . 
&
han \hlc[6d]{sen} \hlc[6l]{guaranteed} \hlc[6d]{safe} \hlc[6l]{return} to \hlc[6l]{cambodia for} all opponents but his strongest \hlc[6l]{critic} \hlc[6d]{, sam rainsy ,} \hlc[6l]{remained} wary . 
\\
\hline
diplomatic efforts to revive the stalled talks appeared to bear fruit monday as japanese foreign affairs secretary \hlc[7d]{of state} nobutaka machimura said \hlc[7d]{king norodom sihanouk} has called on ranariddh and sam rainsy to return to cambodia . 

king norodom sihanouk on tuesday \hlc[7l]{praised agreements} by cambodia 's top two political parties -- previously bitter rivals -- to form a coalition government led by strongman hun sen . 
&
chief \hlc[7d]{of state king norodom sihanouk} \hlc[7l]{praised} the \hlc[7l]{agreement} . 
\\
\hline
\end{tabular}

\end{fontppl}
\end{scriptsize}
\caption{Sample of our ground-truth labels for singleton/pair instances from DUC-04, a multi-document dataset. Ground-truth sentences are widely dispersed among all ten documents. 
}
\label{tab:duc_labels}
\vspace{-0.1in}
\end{table*}

\section{Example Summaries}

Table~\ref{tab:results_output} presents example system summaries and human-written abstracts from CNN/Daily Mail. Each Human Abstract sentence is matched with a sentence singleton or pair from the source document; these singletons/pairs make up the GT-SingPairMix summary. Similarly, each sentence from BERT-Abs is created by compressing a singleton or merging a pair selected by BERT-Extr.

\begin{table*}[t]
\setlength{\tabcolsep}{5pt}
\renewcommand{\arraystretch}{1.1}
\begin{scriptsize}
\begin{fontppl}

\begin{minipage}[b]{0.5\hsize}\centering
\begin{tabular}[t]{|p{2.99in}|}
\hline
\textbf{Extractive Upper Bound}\\[2mm]
\textbullet\, She's a high school freshman with Down syndrome. 
\textbullet\, Trey -- a star on Eastern High School's basketball team in Louisville, Kentucky, who's headed to play college ball next year at Ball State -- was originally going to take his girlfriend to Eastern's prom. 
\\[2mm]

\textbullet\, Trina Helson, a teacher at Eastern, alerted the school's newspaper staff to the prom-posal and posted photos of Trey and Ellie on Twitter that have gone viral.  \\[1mm]

\hline
\hline
\textbf{BERT-Extractive}\\[2mm]

\textbullet\, But all that changed Thursday when Trey asked Ellie to be his prom date. 
\textbullet\, Trey -- a star on Eastern High School's basketball team in Louisville, Kentucky, who's headed to play college ball next year at Ball State -- was originally going to take his girlfriend to Eastern's prom. 
\\[2mm]

\textbullet\, Trina Helson, a teacher at Eastern, alerted the school's newspaper staff to the prom-posal and posted photos of Trey and Ellie on Twitter that have gone viral. 
\\[2mm]

\textbullet\, (CNN) He's a blue chip college basketball recruit. 
\textbullet\, She's a high school freshman with Down syndrome. 
\\[1mm]

\hline
\end{tabular}
\end{minipage}
\hfill
\begin{minipage}[b]{0.5\hsize}\centering
\begin{tabular}[t]{|p{2.99in}|}
\hline
\textbf{Human Abstract}\\[2mm]
\textbullet\, College-bound basketball star asks girl with Down syndrome to high school prom.
 \\[7.8mm]                        
\textbullet\, Pictures of the two during the "prom-posal" have gone viral.
 \\[6.4mm]

\hline
\hline
\textbf{BERT-Abstractive}\\[2mm]
\textbullet\, Trey asked Ellie to be his prom date. \\[13mm]

\textbullet\, Trina Helson, a teacher at Eastern, alerted the school's newspaper staff. \\[5mm]

\textbullet\, He's a high school student with Down syndrome.
 \\[4mm]

\hline
\end{tabular}
\end{minipage}

\vspace{7mm}

\begin{minipage}[b]{0.5\hsize}\centering
\begin{tabular}[t]{|p{2.99in}|}
\hline

\textbf{Extractive Upper Bound}\\[2mm]
\textbullet\, Marseille prosecutor Brice Robin told CNN that "so far no videos were used in the crash investigation." \\[2mm]

\textbullet\, Reichelt told "Erin Burnett: outfront" that he had watched the video and stood by the report, saying Bild and Paris Match are "very confident" that the clip is real. \\[2mm]

\textbullet\, Lubitz told his Lufthansa flight training school in 2009 that he had a "previous episode of severe depression," the airline said Tuesday.\\[2mm]

\hline
\hline
\textbf{BERT-Extractive}\\[2mm]

\textbullet\, Marseille, France (CNN) - the French prosecutor leading an investigation into the crash of Germanwings flight 9525 insisted Wednesday that he was not aware of any video footage from on board the plane. 
\textbullet\, Marseille prosecutor Brice Robin told CNN that "so far no videos were used in the crash investigation." 
\\[2mm]

\textbullet\, Robin's comments follow claims by two magazines, German Daily Bild and French Paris Match, of a cell phone video showing the harrowing final seconds from on board Germanwings flight 9525 as it crashed into the French Alps. 
\textbullet\, The two publications described the supposed video, but did not post it on their websites. 
\\[2mm]

\hline
\end{tabular}
\end{minipage}
\hfill
\begin{minipage}[b]{0.5\hsize}\centering
\begin{tabular}[t]{|p{2.99in}|}
\hline
\textbf{Human Abstract}\\[2mm]   
\textbullet\, Marseille prosecutor says "so far no videos were used in the crash investigation" despite media reports. \\[2mm]                   
\textbullet\, Journalists at Bild and Paris Match are "very confident" the video clip is real, an editor says. \\[4.7mm]
\textbullet\, Andreas Lubitz had informed his Lufthansa training school of an episode of severe depression, airline says. \\[2mm]

\hline
\hline
\textbf{BERT-Abstractive}\\[2mm]
\textbullet\, New : French prosecutor says he was not aware of video footage from on board the plane. \\[10mm]

\textbullet\, Two magazines, including German Daily Bild, have been described as the video. \\[11mm]

\hline
\end{tabular}
\end{minipage}

\end{fontppl}
\end{scriptsize}
\caption{Example system summaries and human-written abstracts. Each Human Abstract sentence is lined up horizontally with its corresponding ground-truth instance, which is found in Extractive Upper Bound summary. Similarly, each sentence from BERT-Abstractive is lined up horizontally with its corresponding instance selected by BERT-Extractive. The sentences are manually de-tokenized for readability.}
\label{tab:results_output}
\vspace{-0.15in}
\end{table*}

\end{document}